\documentclass{article}
\usepackage{spconf,amsmath,graphicx}
\usepackage{algorithm}
\usepackage{algorithmic}
\usepackage{graphics}
\usepackage{caption}
\usepackage{subcaption}
 \usepackage{amssymb} 

\title{Efficient unimodality test in clustering by Signature Testing}
\name{Mahdi Shahbaba and Soosan Beheshti}
\address{Ryerson University\\Department of Computer Science and Electrical Engineering\\350 Victoria Street Toronto, ON M5B 2K3\\mshahbab@ryerson.ca, soosan@ee.ryerson.ca}
\begin{document}
%
\maketitle
\begin{abstract}
This paper provides a new unimodality test with application in hierarchical clustering methods. The proposed method denoted by signature test (Sigtest), transforms the data based on its statistics. The transformed data has much smaller variation compared to the original data and can be evaluated in a simple proposed unimodality test. Compared with the existing unimodality tests, Sigtest is more accurate in detecting the overlapped clusters and has a much less computational complexity.     
Simulation results demonstrate the efficiency of this statistic test for both real and synthetic data sets.
\end{abstract}
\section{Introduction}
\label{sec:intro}
Data clustering is an unsupervised learning algorithm for grouping similar data samples \cite{xu2005survey}. The family of clustering methods only rely on data itself when \textit{a priori} knowledge about the labels and classes is not available. One main challenge in this type of clustering is finding the correct number of clusters involved \cite{chiang2010intelligent}. Hierarchical clustering algorithms answer this problem by using a cluster splitting criteria. These methods test a null hypothesis for distribution of a single cluster and split the dataset until all estimated clusters pass the test. An improper statistical test for splitting criterion will lead to an incorrect estimation of the number of clusters. This problem generally is caused due to the lack of a universal statistic test for all types of clusters. Statistical tests in these approaches are in form of unimodality test. Examples of these unimodality tests are Anderson-Darling\cite{stephens1974edf}, Kolmogorov-Smirnov\cite{lilliefors1967kolmogorov} and dip test\cite{hartigan1985dip}.  

In this paper, we provide a new splitting criterion for unimodality that can also be used in hierarchical clustering algorithms. Our proposed criterion relies on compressing the data based on its statistics and leads to minimizing the data variation by transforming the data. The transformation is denoted in form of signatures. The data signatures for statistical data plays analogous role to the data sparse transformation for sparse signals, i.e., it transforms the data such that the signatures of statistics will be extracted from the data itself. One of the advantages of this statistical test, denoted by Sigtest, is its robustness for recognizing the highly overlapped clusters compared with the state of the art unimodality tests. 
\section{Related Work}
\label{sec:related work}
The correct number of clusters is a crucial parameter which either is available before clustering or should be estimated by clustering methods. X-means algorithm is one of the first hierarchical clustering methods which relies on Bayesian Information criterion (BIC) for cluster splitting \cite{citeulike:305163}, \cite{shahbaba2012improving}. This method only recognizes spherical Gaussian clusters and splits clusters with non-spherical distribution. 

G-means benefits from Anderson-Darling statistic test (AD) for examining the Gaussianity of clusters and similar to X-means it is a wrapper around K-means algorithm. In contrast to X-means, G-means can deal with any distribution from Gaussian family \cite{kinkmeans}, \cite{hu2003comparative}. Employing the Expectation Maximization algorithm (EM), PG-means clustering can deal with overlapped clusters better than G-means \cite{PG-means}. PG-means projects model and all of the dataset on several random directions, and then using Kolmogorov-Smirnov (KS) decides whether model and dataset are matched for each projection. 

Dip-means clustering is constructed based on the Hartigan's dip test of unimodality \cite{kalogeratos2012dip}. According to this clustering method, each sample is a viewer with different distance values from other samples. Using dip test, distribution of the distance values should be examined for unimodality. If all viewers pass the unimodality test then null hypothesis of having a single cluster will be approved. Otherwise, a model with more than one cluster should be considered for the samples. This method is a wrapper around K-means, which can also work with kernel K-means to detect arbitrary shape clusters.    

Both dip-means and G-means rely on statistical tests for cluster splitting, but accuracy of these criteria remains a concern for the case of overlapped clusters. Our proposed statistic test defines probabilistic bounds on signature of a single cluster and employs it as a reference for comparison with any given cluster.
The accuracy of proposed Sigtest will be compared with dip-test, KS and AD using synthetic dataset. Also modified versions of dip-means and G-means based on Sigtest are evaluated using real benchmark datasets from UCI repository database.      
\section{Problem Statement}
\label{sec:problem statement} 
Let $y=[y_{1},y_{2},\cdots,y_{N}]^{T}$ be a vector of $n$ observations; where $y_{i}\in R$ is generated from an unknown distribution.  
There, exist the following possible hypotheses for unimodality test on $y$:\\
$ \hspace{0.5cm}  \bullet  H_{0}$: $y$ is sampled from a unimodal distribution. \\
$ \hspace{0.5cm}  \bullet   H_{1}$: $y$ is not sampled from a unimodal distribution. \\
where, acceptance of each hypothesis will affect the cluster splitting criterion of the model. 
\subsection{Application of Unimodality Test in Clustering}
\begin{figure}[h]
\centerline{\includegraphics[trim = 10mm 10mm 8mm 10mm, clip, scale=0.65,width=5cm]{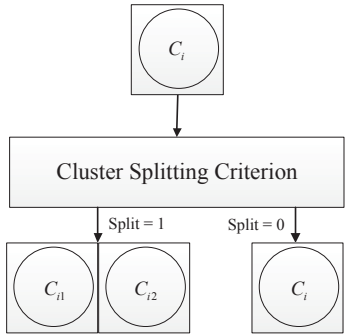}}
\caption{Hierarchical clustering and data splitting.}
\label{HC}
\end{figure}

Hierarchical clustering methods rely on cluster splitting criteria to recognize single clusters in a given dataset. In Fig.\ref{HC}, $C_{i}$ represents the data which should be checked for splitting at the $i^{th}$ stage of clustering. If the criterion accepts splitting ($H_{1}\equiv split=1$), $C_{i}$ will be split into two new clusters $C_{i1}$ and $C_{i2}$, otherwise ($H_{0}\equiv split=0$) it remains as one cluster. In the following stages, the checking procedure continues for all new clusters (if any) until $H_{0}$ is valid for all clusters. 
\\In this clustering, the available data ($x$) has dimension of $d$, however, splitting criterion is usually based on a transformation of $d$-dimensional data to a $one$-dimensional data. The first step in using the criterion is transforming $x$ to $y$:
\begin{eqnarray}
y = f(x)
\end{eqnarray}
where $f$ is the transformation of $x$ to $one$-dimensional $y$.     
\section{Unimodality Signature Test (Sigtest)}
\label{sec:method}
\begin{figure}[!h]
\centerline{\includegraphics[trim = 15mm 19mm 10mm 5mm, clip, width=9cm]{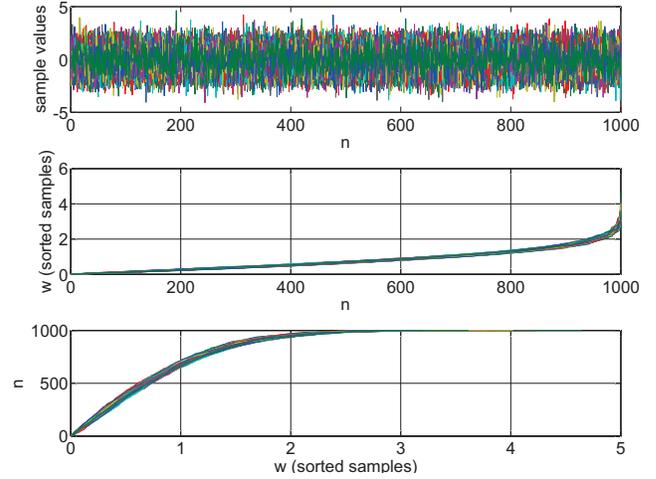}}
\caption{Top figure: 100 runs of a zero mean and unit variance Gaussian distribution with length 1000. Middle: sorted absolute values of the top figure. Bottom: The same middle figure with swapped axes.}
\label{signature}
\end{figure}

In this section, we define a new unimodality test for hierarchical clustering which relies on data signature and probabilistic bounds on its distribution. Signature of a data is a function of data that compresses the data in a proper transformation\cite{Nide}. To illustrate an example of data signature, Fig.\ref{signature} shows 1000 randomly generated Gaussian samples from $\mathcal{N}(0,1)$ for 100 runs. As the figure illustrates, while samples themselves (in the top plot) vary between $\pm3\sigma$, the middle plot which is sorted version of the same samples is a transformation of the top plot in a much more compact form, i.e., the variance of the sorted version is smaller than $\frac{1}{10}$ of the original variance.
Therefore, we can define a probabilistic confidence region around the dense area, with a very small variance, and use it as a signature for unimodality. 
\subsection{Probabilistic Bounds on Signature of Unimodal Distribution}
In the following, we propose two signatures $g_{1}(w_{n})$ and $g_{2}(w_{n})$ for unimodality test. Let $w =[w_{1},w_{2},\cdots,w_{N}]$ be sorted absolute values of samples from a unimodal distribution. The first suggested signature as it was shown in Fig.\ref{signature} can be the sorted version of data itself:
\begin{eqnarray}
\label{sig1}
 g_{1}(w_{n}) = w_{n}
\end{eqnarray}
This signature has the following expected value and variance: 
\begin{eqnarray}
\label{E[w_{n}]}
E[w_{n}] &=& F(w_{n})\\
\label{var[w_{n}]}
var[w_{n}] &=& \frac{1}{N} F(w_{n}) (1-F(w_{n}))
\end{eqnarray}
where $F(w_{n})$ is the cumulative distribution function (cdf) of $w_{n}$.
Details of calculation of this expected value and variance are provided in \cite{Nide}.
Another proposed signature $g_{2}(w_{n})$ is:
\begin{equation}
\label{sig2}
 \hspace{0.2cm} g_{2}(w_{n}) = \frac{1}{n}\sum_{j=1}^{n} w_{j}^{}\\
\end{equation}    
Using (\ref{E[w_{n}]}), the expected value of $g_{2}(w_{n})$ is:
\begin{equation}
\label{E2}
E[g_{2}(w_{n})] = E[w_{n}]
\end{equation}
and using (\ref{var[w_{n}]}), it can be shown that the variance of $g_{2}(w_{n})$ is bounded as follows\footnote{According to the Cauchy-Schwarz inequality we have:
\begin{equation}
cov(w_{k},w_{l}) \le  \sqrt{var[w_{k}]} \sqrt{var[w_{l}]} 
\end{equation}
consequently:
\begin{eqnarray}
\label{var2}
var[g_{2}(w_{n})] = \frac{1}{n^{2}}(\sum_{j=1}^{n} var[w_{j}] 
+\sum_{ k\ne l \ge 1}^{n} cov(w_{k},w_{l}) \nonumber \\
\le \frac{n}{n^{2}} var[w_{n}] + \frac{2}{n^2}\frac{n(n-1)}{2} var[w_{n}]
\end{eqnarray}
}:
\begin{equation}
var[g_{2}(w_{n})] \le var[w_{n}]
\end{equation}
Note that both of the above signatures have very small variances compared to their original distribution similar to what is shown in Fig.\ref{signature}.  
The tight boundaries of each signature as a function of their indexes are denoted by $U(n)$ and $L(n)$: 
\begin{eqnarray}
\label{UL}
U(n) =E[g_{i}(w_{n})]+\gamma \sqrt{var[g_{i}(w_{n})]}\\
L(n) = E[g_{i}(w_{n})]-\gamma \sqrt{var[g_{i}(w_{n})]} \nonumber
\end{eqnarray}
where $\gamma$ is chosen based on the desired confidence probability. For example, $\gamma =2\sigma$ gives $95\%$ confidence probability for the Gaussian distribution.  
\subsection{Sigtest for The Available Data}
In the following we show how the proposed signature boundaries in (\ref{UL}) can be used for the unimodality test. Let $z =[z_{1},z_{2},\cdots,z_{N}]$ be sorted absolute values of available data $y$, which its unimodality is unknown. Using signatures in (\ref{sig1}) and (\ref{sig2}), we can define our signature tests (Sigtest) as following:
\begin{eqnarray}
\label{sigz1}
Sigtest_{1}&:& \hspace{0.2cm}  g_{1}(z_{n}) = z_{n}\\
\label{sigz2}
Sigtest_{2}&:& \hspace{0.2cm} g_{2}(z_{n}) = \frac{1}{n}\sum_{j=1}^{n} z_{j}^{}
\end{eqnarray} 

To test the unimodality of $z$, $g_{i}(z_{n})$ should be compared with the probabilistic bounds (based on our defined signatures, here $i$ can be 1 or 2): 
\begin{equation}
c_{n} = 
\begin{cases}
    0,& L(n)<g_{i}(z_{n})<U(n) \\
    1,& otherwise  
\end{cases}
\end{equation}
\begin{equation}
C = \frac{1}{N}\sum_{n=1}^{N}c_{n} 
\end{equation} 
where $c_{n}$ shows any mismatch between the bounds and the signature at index $n$, and $C$ is the total counting index. Consequently, the test chooses one of the hypotheses in Section \ref{sec:problem statement} based on the following comparison:
\begin{equation}
\label{T}
C\quad\mathop{\gtrless}_{H_{0}}^{H_{1}}\quad T_{}
\end{equation}
where $T_{}$ is the threshold for a chosen confidence probability. Algorithm \ref{algorithm} demonstrates steps of the Sigtest. 
\begin{algorithm}
\caption{Unimodality Signature Test}
\label{algorithm}
\begin{algorithmic} [1]
{\small
\REQUIRE input samples $x=\{x_{i}\}_{i=1}^{N}$, $x_{i}\in R^{d}$, threshold $T$. \\
\ENSURE  result of the splitting test, split = $0$ or $1$. \\
\renewcommand{\algorithmicrequire}{\textbf{Input:}}
\renewcommand{\algorithmicensure}{\textbf{Output:}}
\newcommand{\algorithmicbreak}{\textbf{break}}
\newcommand{\BREAK}{\STATE \algorithmicbreak}
\STATE $C \leftarrow 0 $ 
\STATE $y \leftarrow f(x) $ 
\STATE $ g_{1}(z_{n}) \leftarrow sort(abs(normalize(y)))$ 
\STATE $ g_{2}(z_{n}) \leftarrow cumsum(g_{1}(z_{n}))$ 
\STATE $ compute \hspace{0.1cm} U(n)\hspace{0.1cm} and \hspace{0.1cm}L(n)\hspace{0.1cm} from \hspace{0.3cm} (\ref{UL})$ 
\FOR{$j=1$ to $N$}
\IF{$g_{i}(z_{j}) > U(j) \hspace{0.3cm}or\hspace{0.3cm} g_{i}(z_{j}) < L(j) $}
\STATE $C \leftarrow C  + 1$
\ENDIF
\ENDFOR 
\IF{$C > T $}
\STATE $split \leftarrow 1$
\ELSE  
\STATE {$split \leftarrow 0$}
\ENDIF
}
\end{algorithmic}
\end{algorithm}

\begin{figure}[htb]
\begin{minipage}[b]{0.48\linewidth}
  \centering
  \centerline{\includegraphics[trim = 25mm 33mm 25mm 20mm, clip,width=4cm]{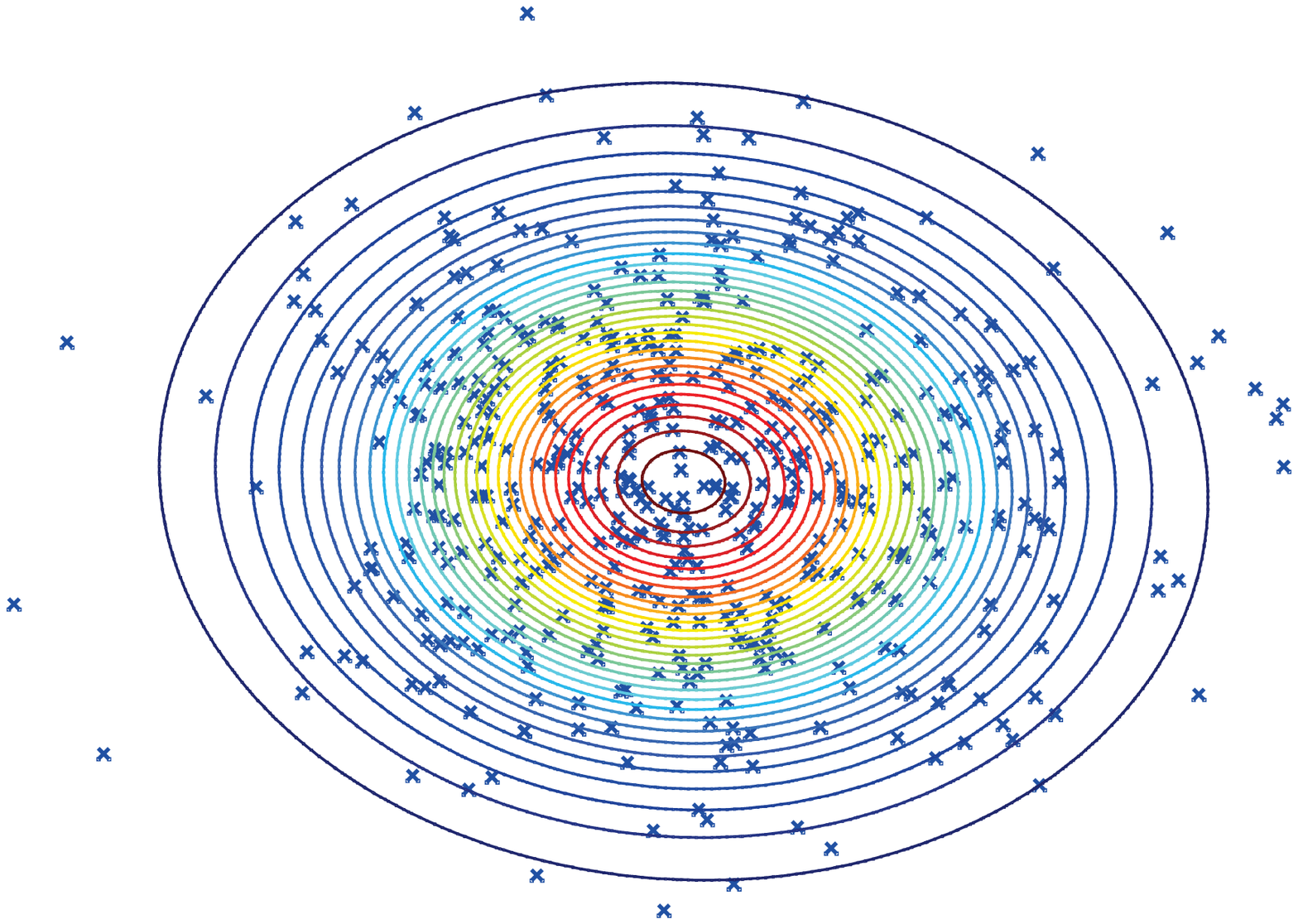}}
  \centerline{(a) }\medskip
\end{minipage}
\begin{minipage}[b]{0.48\linewidth}
  \centering
  \centerline{\includegraphics[trim = 20mm 15mm 25mm 20mm, clip,width=4cm]{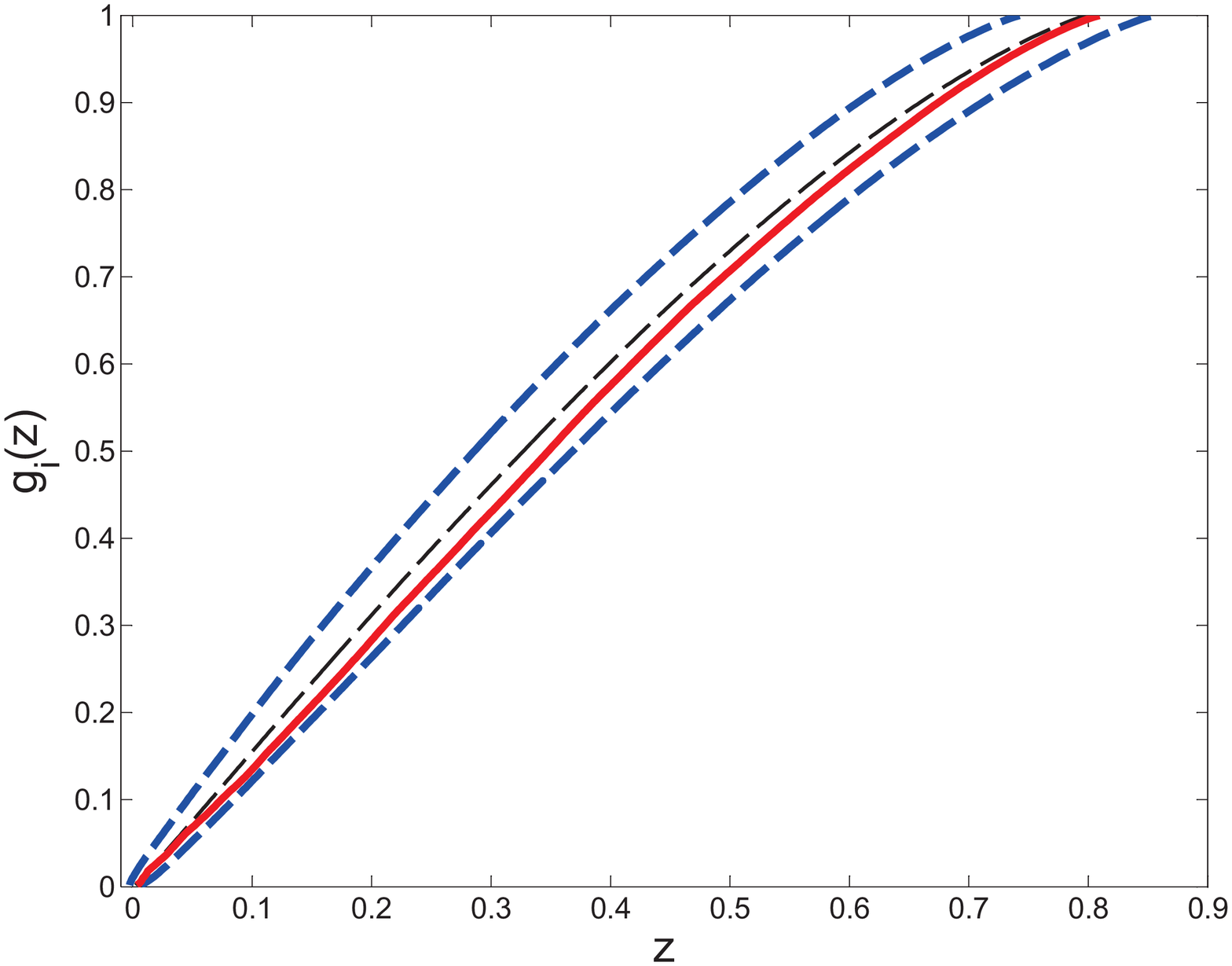}}
  \centerline{(b)}\medskip
\end{minipage}
\begin{minipage}[b]{.48\linewidth}
  \centering
  \centerline{\includegraphics[trim = 25mm 33mm 25mm 20mm, clip,width=4.0cm]{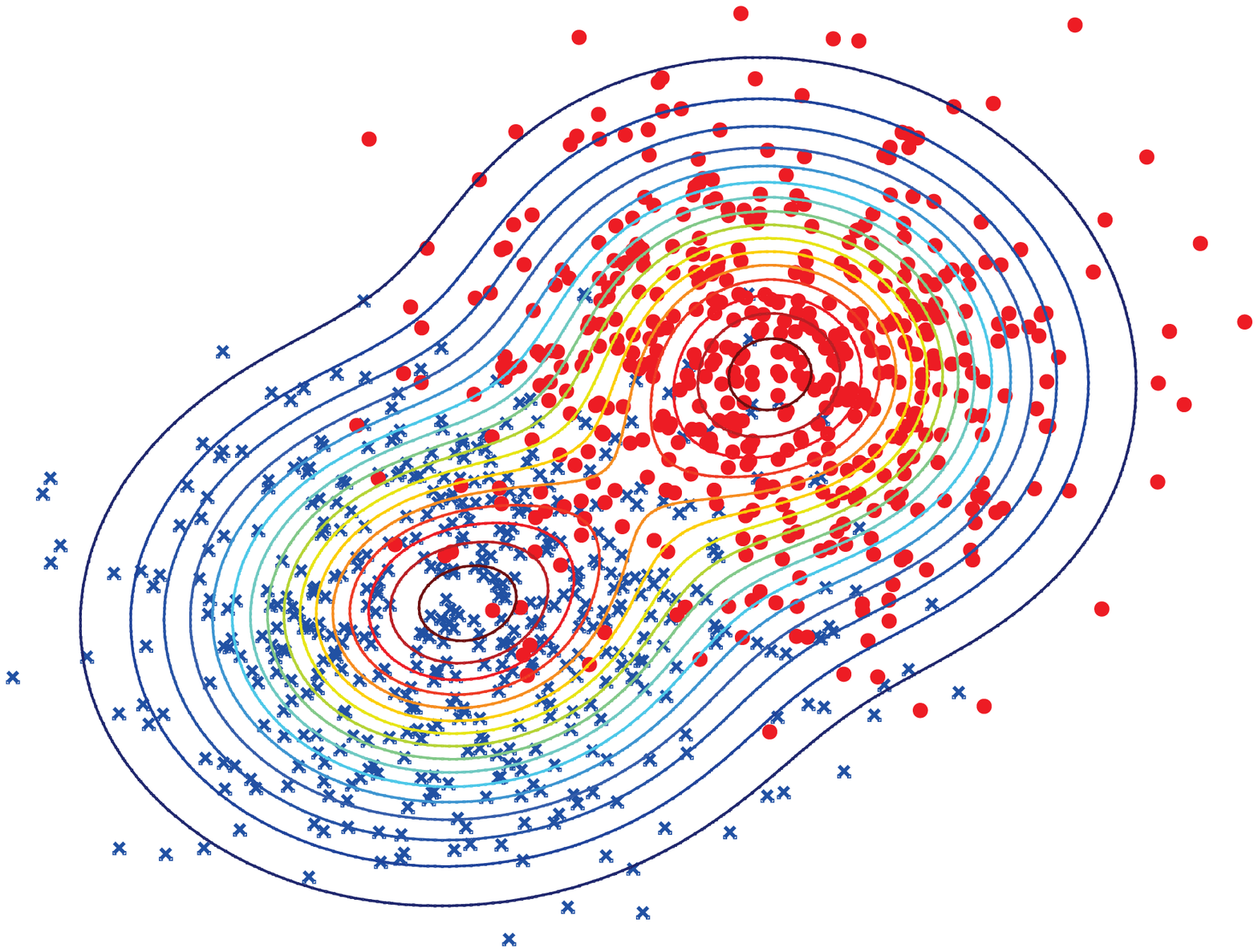}}
  \centerline{(c)}\medskip
\end{minipage}
\hfill
\begin{minipage}[b]{0.48\linewidth}
  \centering
  \centerline{\includegraphics[trim = 25mm 20mm 25mm 20mm,width=4.0cm]{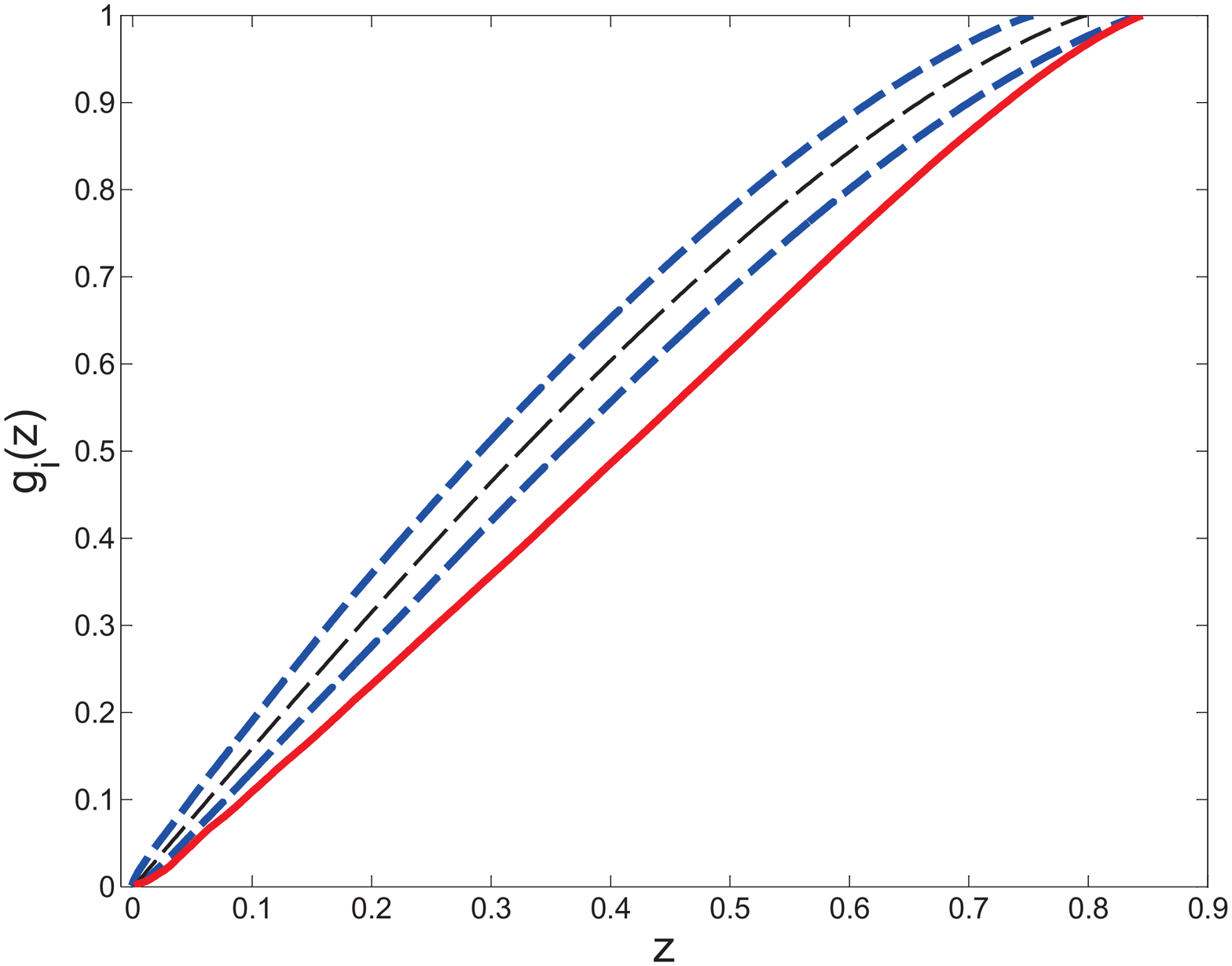}}
  \centerline{(d)}\medskip
\end{minipage}
\begin{minipage}[b]{.48\linewidth}
  \centering
  \centerline{\includegraphics[trim = 25mm 33mm 25mm 20mm,width=4.0cm]{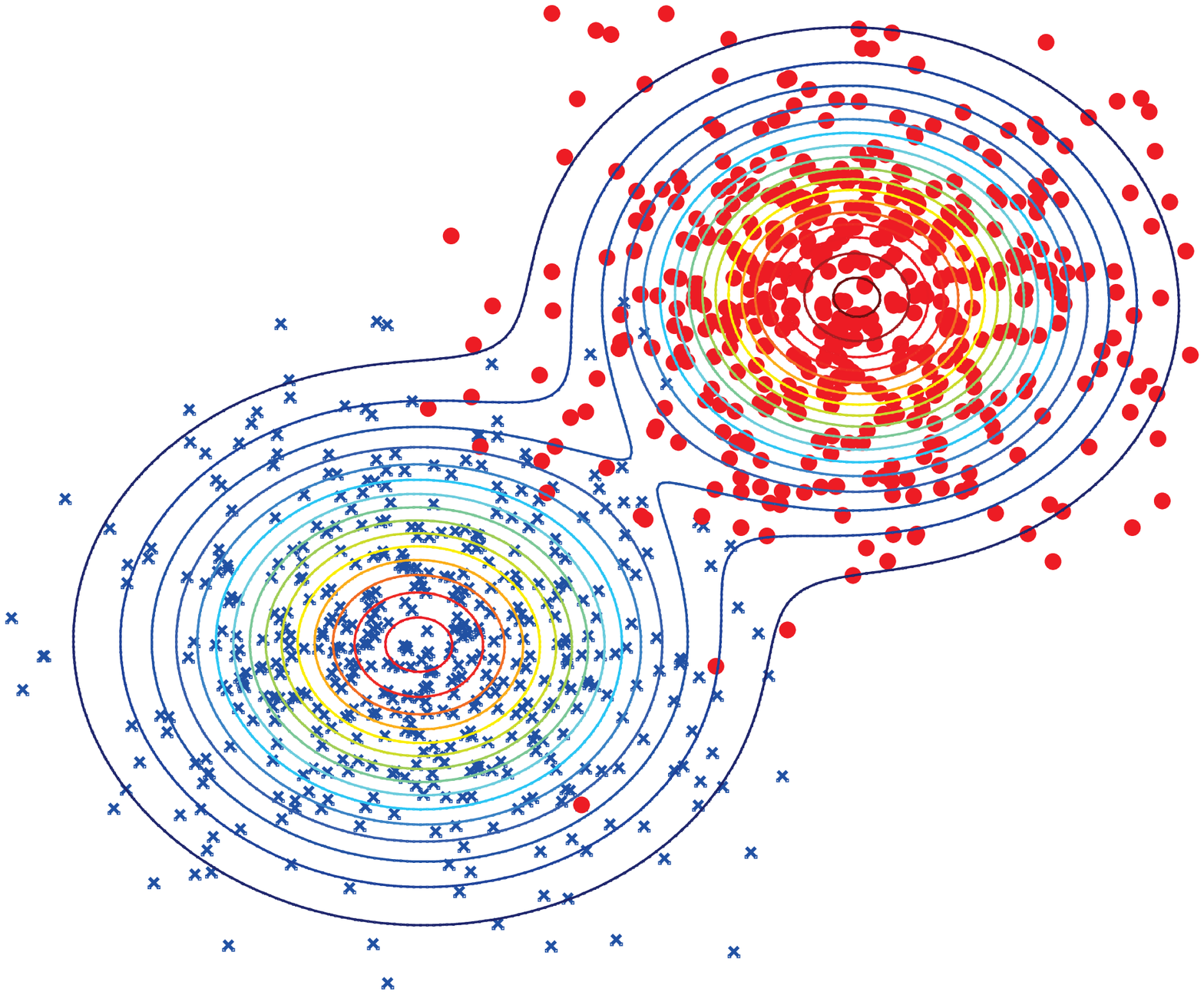}}
  \centerline{(e) }\medskip
\end{minipage}
\hfill
\begin{minipage}[b]{0.48\linewidth}
  \centering
  \centerline{\includegraphics[trim = 25mm 20mm 25mm 20mm,width=4.0cm]{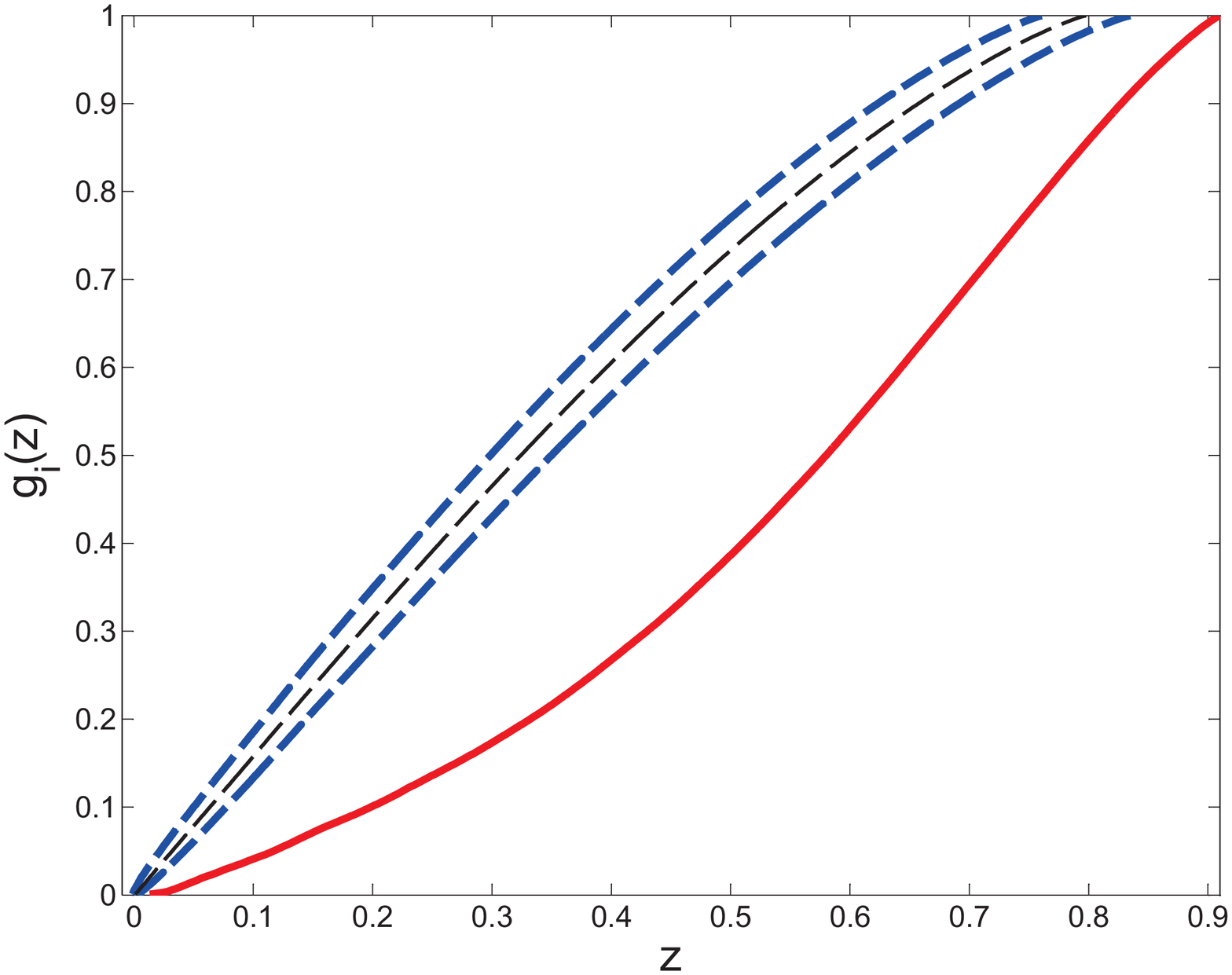}}
  \centerline{(f)}\medskip
\end{minipage}
\caption{Sigtest for single (a), and overlapped clusters (c) and (e). The two clusters (c) and (e) are separated with $2\sigma$ and $3\sigma$ respectively.}
\label{fig:res}
\end{figure}

The behavior of Sigtest for 95$\%$ confidence probability (T=0.4) on synthetic clusters is shown in Fig.\ref{fig:res}. The left figures are clusters (a single and two overlapped clusters). The  right figures are the behavior of the associated Sigtest. Bounds of the unimodal distribution ($U(n)$ and $L(n)$) are in blue dashed lines, while the test data $z$ is the red line. The Sigtest of the first cluster lies completely inside the boundaries ($C=0$), while for both (c) and (e) Sigtest is out of the boundary test which results in large values for $C$ (0.95 and 0.99). Therefore, the method splits clusters in (c) and (e).

\subsection{Role of the Signatures}
 In this paper we have proposed two signatures. While the first signature $g_{1}(w_{n})$ deals with unimodal Gaussian distributions, the second signature $g_{2}(w_{n})$ can work with any distribution from the unimodal family due to the weak law of large numbers, as the summation in (\ref{sig2}) converges to Gaussian distribution for a large length of $n$. Consequently, even if data samples are non-Gaussian, $g_{2}(w_{n})$ behaves similar to Gaussian. 

\section{Simulation Results}
\label{sec:results}
In the first set of simulations, AD, KS, dip test, and Sigtest were used to examine the unimodality of two overlapped Gaussian clusters. Each cluster has $100$ samples and its variance is $\sigma^{2}=1$. Table.\ref{statistic_tests} shows the success rate of each statistic test when clusters are overlapped with different distances. The default significant levels are $0.0001$ for AD, zero for Dip test, and $0.05$ for KS. With $95\%$ confidence probability in Sigtest, $\gamma$ and $T$ in (\ref{UL}) and (\ref{T}) are $2$ and 0.4 respectively. As the table shows the optimum methods (specially with more overlapping) are the Sigtests. In addition, the low computational complexity of the Sigtests resulted in much smaller computation time.   
\begin{center}
 \begin{table}[!th]
\renewcommand{\arraystretch}{1.4}
\caption{Success rates of statistic tests for detecting two overlapped clusters with different central distances (averaged over 100 runs).}
\label{statistic_tests}
\centering
\scalebox{0.68}{
 \hfill{}
\begin{tabular}{ c  c  c  c c c  c}
\hline
&  \multicolumn{5}{c}{Distance between center of clusters }           \\
 \cline{2-6} 
Tests &   $2\sigma$   & $2.25\sigma$  & $2.5\sigma$   & $2.8\sigma$ & $3\sigma$ & Average time (s)     \\
\hline
$Sigtest_{2}$(\%)    &56&93& 99 &100&100&$0.2\times10^{-4}$\\
\hline
$Sigtest_{1}$(\%)    &69&97 &100 &100&100&$0.2\times10^{-4}$\\
\hline
AD(\%)    &29&76 &97 &100&100&$3.96\times10^{-4}$\\
\hline
KS (\%)   &10& 37&74 &95&100&$30\times10^{-4}$\\
\hline
dip (\%)   &3& 8 &21&82&94&$2197\times10^{-4}$\\
\hline
\end{tabular}
 \hfill{}
}
\end{table}
 \end{center}
\subsection{Application in Clustering}
We denote dip-means and G-means when their splitting criteria are replaced with Sigtest as G-means$^{+}$ and dip-means$^{+}$. The comparison results on benchmark datasets are presented in Table \ref{table2}. The quality of clustering is examined by Variation of Information (VI) \cite{meilua2007comparing} and Adjusted Rand Index (ARI) \cite{hubert1985comparing}, where smaller VI and larger ARI are desired. Here, $m^{*}$ is the correct number of clusters, and $d$ is dimension of the data. Results are given for an average over 20 simulations. As the table shows G-means$^{+}$ and dip-means$^{+}$ consistently perform better than dip-means and G-means.  
\begin{center} 
\begin{table}[h!]
\renewcommand{\arraystretch}{1}
\caption{Comparison between G-means, dip-means and their improved version.}
\label{table2}
\centering
\scalebox{0.7}{
 \begin{tabular}{ l c c c c}
\hline
Data set &   G-means   & G-means$^{+}$ & dip-means  &  dip-means $^{+}$          
\tabularnewline
\hline
Iris&4.5$\pm$0.50&3$\pm$0&2$\pm$0&2.6$\pm$0.49\\
$(m^{*}=3, d=4)$&\\
VI&0.84$\pm$0.11&0.68$\pm$0.13&0.64$\pm$0.11&0.60$\pm$2.29\\
ARI&0.53$\pm$0.07&0.58$\pm$0.14&0.53$\pm$4.48&0.56$\pm$0.11\\
\hline 
Optical digits&25.8$\pm$3.34&14.6$\pm$1.51&1$\pm$0&6.2$\pm$1.64\\
$(m^{*}=10, d=64)$\\
VI&1.31$\pm$0.08&1.14$\pm$0.10&2.3025$\pm$0&1.76$\pm$0.09\\
ARI&0.57$\pm$0.03&0.66$\pm$0.04&0$\pm$0&0.35$\pm$0.05\\
\hline
Leukemia&4$\pm$0&3$\pm$0&1.75$\pm$0.44&3.1$\pm$0.41\\
$(m^{*}=3, d=39)$\\
VI&0.49$\pm$0.00&0.30$\pm$0&0.84$\pm$0.14&0.67$\pm$0.18\\
ARI&0.77$\pm$0.00&0.88$\pm$0&0.39$\pm$0.23&0.59$\pm$0.14\\
\hline 
Seed &4$\pm$0&2$\pm$0.72&1$\pm$0&3$\pm$0\\
$(m^{*}=3, d=7)$\\
VI&0.87$\pm$0.00&0.84$\pm$0.15&1.0986$\pm$0&0.66$\pm$0\\
ARI&0.41$\pm$0.26&0.61$\pm$0.00&0$\pm$0&0.71$\pm$0\\
\hline
Pendigits &77.2$\pm$2.49&24.4$\pm$3.20&7$\pm$0&10.2$\pm$0.44\\
$(m^{*}=10, d=16)$\\
VI&2.01$\pm$0.03&1.38$\pm$0.01&1.5866$\pm$0&1.4013$\pm$0.00\\
ARI&0.27$\pm$0.01&0.51$\pm$0.01&0.34$\pm$0&0.57$\pm$0.00\\
\hline
\end{tabular}
}
\end{table}
\end{center}
\section{Conclusions and Future Work}
\label{sec:conclusion}
In this paper, we introduced the idea of using signature test (Sigtest) for cluster splitting criterion. 
The two proposed signatures can compress the data based on its statistics and represent it in a space with smaller variation. 
The advantage of the proposed Sigtest compared to similar methods is in its robustness for recognizing the overlapped clusters, while its complexity is much less than the compared methods.
The simulation results shows that replacing the existing splitting tests with Sigtest in hierarchical clustering improves the accuracy of estimated number of clusters as well as clustering quality. As future work, more signatures can be proposed for a general unimodal distribution or for a specific distribution in splitting criterion.    
%
%
%


\bibliographystyle{IEEEbib}
\bibliography{strings,refs }
\end{document}